\documentclass[11pt]{article}
\PassOptionsToPackage{table}{xcolor}
\usepackage{acl}  
\usepackage{times}
\usepackage{float}
\usepackage{subfig}
\usepackage{latexsym}
\usepackage[T1]{fontenc}
\usepackage[utf8]{inputenc}
\usepackage{microtype}
\usepackage{inconsolata}
\usepackage{graphicx}     
\usepackage{algorithm}    
\usepackage{algpseudocode}
\usepackage{amsmath}      
\usepackage{booktabs}     
\usepackage{multirow}     
\usepackage{float}        
\usepackage{hyperref}     
\usepackage{graphicx}
\usepackage{subfig}
\usepackage{mdframed}      

\usepackage{booktabs}
\usepackage{multirow}
\usepackage{graphicx}
\title{MoE-DisCo:   Low Economy Cost Training Mixture-of-Experts Models}
\author{Xin Ye\textsuperscript{1,2} \and Daning Cheng\textsuperscript{1*} \and Boyang Zhang\textsuperscript{1,2,3} and Yunquan Zhang\textsuperscript{1}  \\
  \textsuperscript{1}Institute of Computing Technology, Chinese Academy of Sciences, Beijing, China \\
  \textsuperscript{2}University of Chinese Academy of Sciences, Beijing, China \\
   \textsuperscript{3}Peng Cheng Laboratory, Shenzhen, China \\
}

\begin{document}
\maketitle  

\begin{abstract}

Training large-scale Mixture-of-Experts (MoE) models typically requires high-memory, high-bandwidth GPUs (e.g., A100), and their high cost has become a major barrier to large-model training. In contrast, affordable hardware is low-cost but constrained by by memory capacity and bandwidth, making it unsuitable for direct LLM training. To address this, we propose MoE-DisCo (Mixture-of-Experts with Disentangled Clustering and Coordination)—a staged training framework. MoE-DisCo decomposes the MoE model into multiple dense submodels, each consisting of a shared backbone and a single expert, and partitions the training data into subsets using unsupervised clustering. Each submodel is trained independently and in parallel on its assigned data subset using low-cost devices, without any inter-device communication. Subsequently, all experts are integrated into a complete MoE model and fine-tuned globally for a short period on high-memory, high-bandwidth GPUs. Experiments show that our method matches or even surpasses full-parameter training in performance across multiple downstream tasks, loss function and perplexity (PPL) with a cost reduction of 47.6\% to 69.5\% on Qwen1.5-MoE-2.7B and Llama-MoE-3.5B across different datasets.
\end{abstract}

\section{Introduction}

Large Language Models (LLMs) and Mixture of Experts (MoE) have significantly advanced the field of natural language processing, yet their training economy cost has become a major barrier to broad participation. Current training paradigms heavily rely on high-memory GPUs such as NVIDIA A100/H800, which can cost over \$2.28 per GPU-hour in cloud environments. In contrast, affordable computing devices, such as  DCUs, Ascend 910A, or consumer-grade GPUs, cost less. For example, the cost of DCU is less than \$0.03 per hour. But, their limited memory capacity (typically $\leq$ 24~GB) and bandwidth make them seemingly unsuitable for training models at the billion- or trillion-parameter scale.

\begin{figure}[!htbp]
  \centering
  \includegraphics[width=0.3\textwidth]{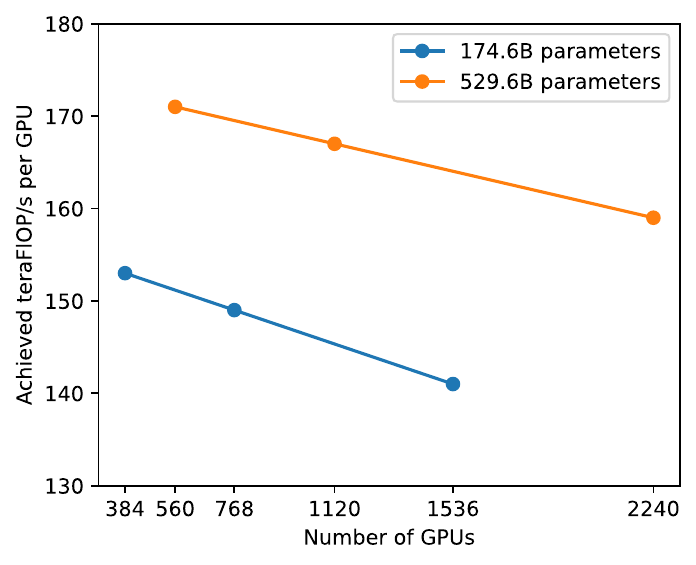} 
   
  \caption{Model FLOPs Utilization (MFU) per GPU under 3D parallelism for language models of different scales as the number of GPUs increases. As the cluster scales from tens to thousands of GPUs, MFU per GPU consistently declines across all model sizes.\cite{DBLP:journals/corr/abs-2104-04473}}
  \label{fig:google data}
\end{figure}

Furthermore, in large-scale model training, as the scale of GPU deployment expands, the per-GPU computational utilization almost inevitably declines. This is because larger clusters introduce more pronounced communication overhead—such as gradient synchronization and activation transmission—along with pipeline bubbles in pipeline parallelism, inter-node bandwidth bottlenecks, and load imbalance. These factors cause each GPU to spend more time waiting rather than computing, shown in Figure \ref{fig:google data}.

\begin{mdframed}[leftline=true, rightline=false, topline=false, bottomline=false, linewidth=2.5pt, linecolor=magenta!50, backgroundcolor=gray!10]
Therefore, using low-cost hardware as much as possible and minimizing multi-GPU coordination is key to reducing the cost of large model training. 
\end{mdframed}

\begin{figure}[!htbp]
  \centering
  \includegraphics[width=0.23\textwidth]{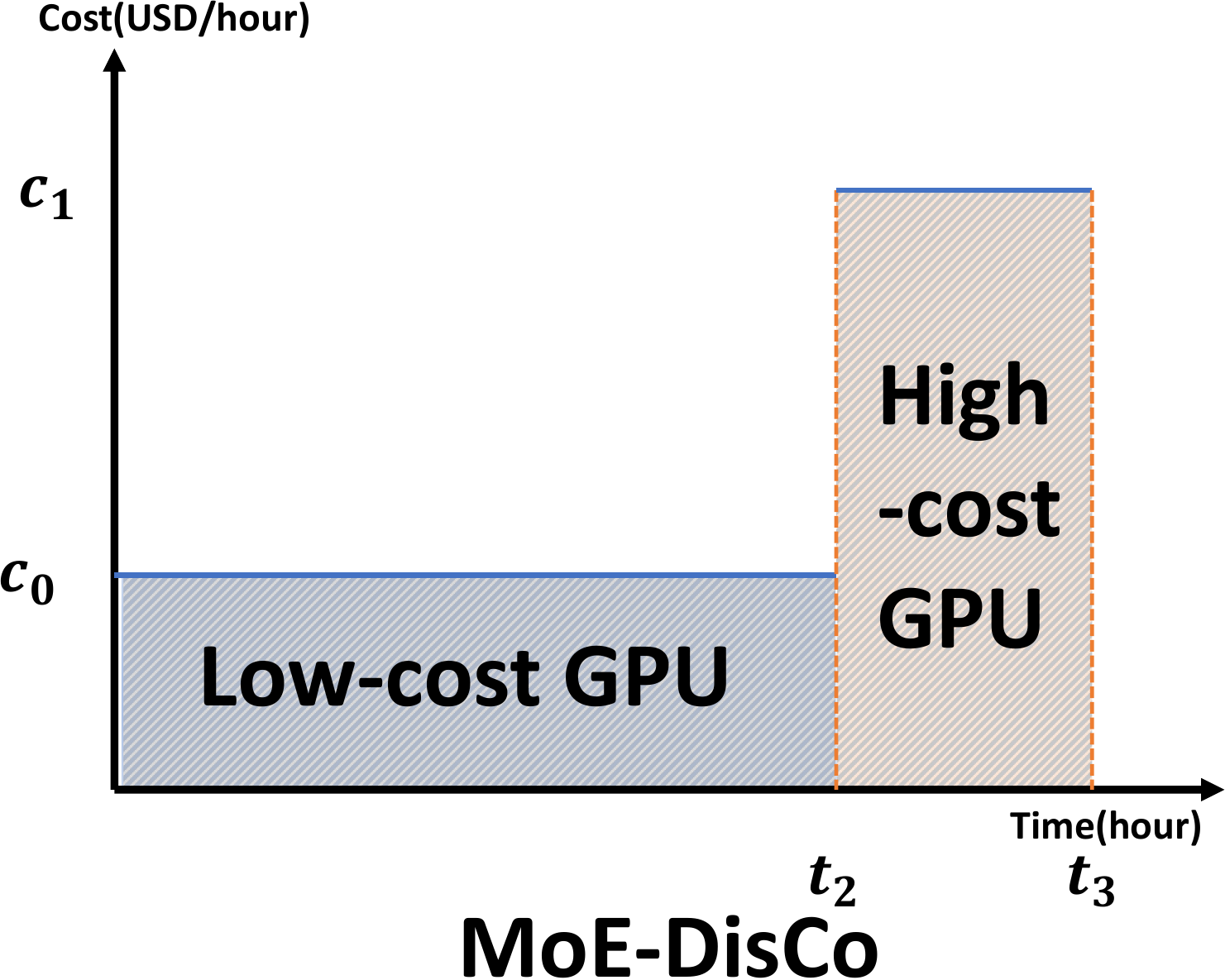} 
  \includegraphics[width=0.23\textwidth]{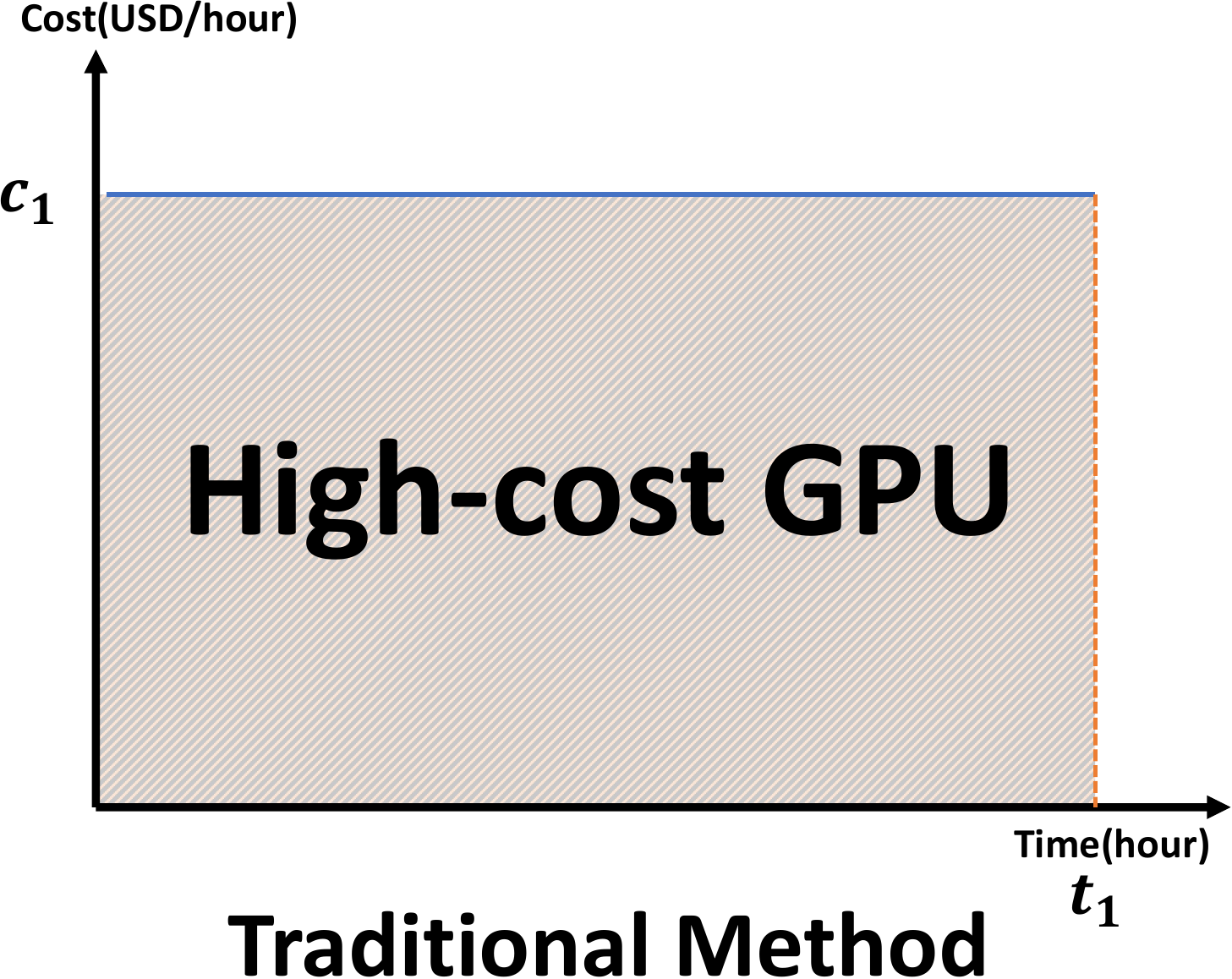}
  \caption{Comparison of training cost profiles between MoE-DisCo (left) and traditional  MoE training (right). MoE-DisCo first performs the majority of training on low-cost GPUs (cost $c_0$), which can be highly parallelized across multiple devices, followed by a short fine-tune phase on high-cost GPUs (cost $c_1$). In contrast, traditional methods rely solely on high-cost hardware throughout the entire training process. The total cost (shaded area under the curve) is significantly reduced in MoE-DisCo, demonstrating its efficiency in lowering the monetary burden of large-scale MoE training.}
  \label{fig:moedisco_pipeline}
\end{figure}

To address high-cost challenge, we propose MoE-DisCo (Mixture-of-Experts with Disentangled Clustering and Coordination), a staged, hardware-aware training framework for MoE models. Specifically, MoE-DisCo decomposes a full MoE model with K-experts for each MoE layers into K independent dense submodels, each consisting of the shared Transformer backbone plus a single expert. Concurrently, we partition the original training data into K subsets using unsupervised clustering, establishing a disentangled alignment between experts and data clusters. Each submodel is trained exclusively on its assigned data subset, without any inter-model communication. Since each submodel is equivalent to a standard dense model (with a parameter scale much smaller than that of the full MoE model), it can be efficiently trained on a single low cost device. Crucially, this parallel training phase eliminates inter-device communication overhead. After all submodels are trained, MoE-DisCo reintegrates the individual expert modules into a unified MoE architecture and performs a brief global fine-tune phase on the full dataset. Although this final stage requires high-memory GPUs, it takes significantly less time than the total training time of conventional approaches, shown in Figure \ref{fig:moedisco_pipeline}. 

Our main contributions are as follows: (1) We propose MoE-DisCo—a low-cost, staged MoE training framework designed for resource-constrained hardware; (2) We show that two stages of MoE-DisCo can be executed on hardware of different cost hardware, significantly reducing the overall training economy cost of large-scale MoE systems; (3) We evaluated MoE-DisCo across multiple datasets and model architectures. Results show that MoE-DisCo effectively reduces the usage of high-cost GPUs while producing MoE models whose accuracy matches or even exceeds that of models trained with the original  MoE approach with a cost reduction of 47.6\% to 69.5\% on Qwen1.5-MoE-2.7B and Llama-MoE-3.5B across different datasets. The code address is shown in Appendix.

\section{Related Work}
 
The Mixture of Experts (MoE) is an architectural paradigm that enhances model capacity and computational efficiency \cite{2014Adaptive} in different areas \cite{2021CPT,2024DB,2024Deep}. Subsequently, the work \cite{2017Outrageously} proposed the sparsely-gated MoE mechanism, which enables significant model scaling under a fixed computational budget—by activating only the top-K experts during each forward pass. GShard \cite{2020GShard} was the first to successfully apply MoE to the Transformer architecture and introduced the notion of “expert capacity”.  Subsequent works have improved MoE from multiple perspectives, Switch Transformer \cite{2021Switch}, Hash FFN, proposed by Facebook AI \cite{2021Hash},  StableMoE \cite{2022StableMoE}. 

The past decade,  the researchers  propose  algorithms and high-performance software to make MoE practically useful \cite{2017GPU, 2017Block,2021Efficient,2021GLaM}. Most of them are working on high performance framework \cite{2020ZeRO,2021Scalable,he2021fastmoe,he2022fastermoe,liu2025netmoe,jin2025megascale,2024MoESys,jin2025megascale,jin2025bigmac,yuan2025x}. There are a few works working on how to train MoE model with low economy cost in algorithm view, which works well with system.

\begin{figure*}[t]
  \centering
  \makebox[\linewidth][c]{
    \includegraphics[width=0.9\linewidth]{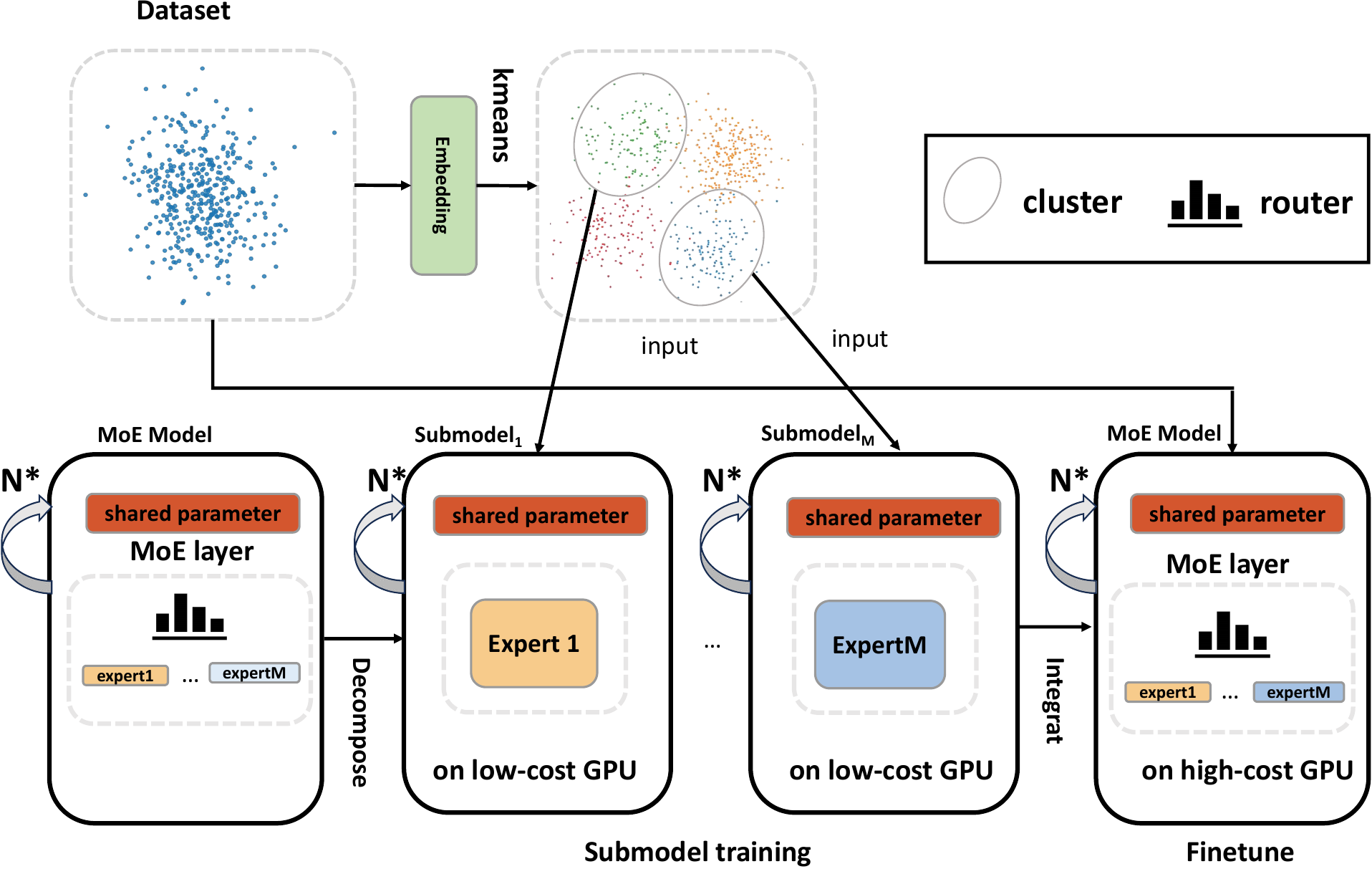}  
  }
  \caption{Overview of the MoE-DisCo training framework. The original MoE model is decomposed into multiple dense submodels, each containing a single expert and shared parameters. Training data is clustered via k-means to assign semantically distinct subsets to individual experts. Submodels are trained independently on low-cost GPUs, enabling high parallelism and reduced hardware cost. Finally, the experts are integrated into a unified MoE architecture and fine-tuned on a high-cost GPU using the full dataset.}
  \label{fig:training_flow}
\end{figure*}

\section{Methodology}
In standard MoE training paradigms, although only a small number of experts are activated during inference, all expert parameters must be loaded and updated simultaneously during training to support backpropagation and gradient updates. This causes memory and computational overhead to grow linearly with the number of experts, severely limiting the trainability of MoE models on low-cost GPUs.

The core question addressed in this paper is: without significantly sacrificing the final performance of the model, can we alter the parameter update strategy to minimize the need for simultaneous loading of all experts during MoE training, thereby drastically reducing reliance on high-cost GPUs? To this end, we introduce Block Coordinate Descent (BCD) and SimulParallel SGD \cite{2011Parallelized} as the core optimization idea for MoE training and design an expert-level block training framework based on the structural characteristics of MoE, enabling the main training stages of large-scale MoE models to be completed on low-memory, low-cost GPUs.

\subsection{Start Point and Theoretical Foundation}




Our approach is inspired by the theoretical foundations of two classical optimization frameworks: Block Coordinate Descent (BCD) and SimulParallel SGD \cite{2011Parallelized}. These provide critical insights into low-cost MoE training from the perspectives of parameter update strategies and distributed training architectures, respectively.

BCD  is an iterative algorithm widely used for large-scale non-convex optimization. Its core idea is to update only one block of parameters per iteration while keeping all others fixed, thereby significantly reducing memory and computational overhead. The MoE
architecture exhibits similar sparsity during inference—the gating network activates only the Top-$K$ experts, avoiding computation over the entire model. However, in standard training, backpropagation must traverse all expert paths, making it impossible to exploit this sparsity for memory savings.

We propose integrating BCD into MoE training: in each step, only one expert and the shared backbone are updated, while all other experts remain frozen. In this setting, each training unit maintains only a dense sub-model equivalent to a single-expert branch. This enables training on low-memory devices. Moreover, different experts can be trained fully in parallel without any communication or synchronization.

However, merely decoupling parameters is insufficient to guarantee performance. The key challenge lies in assigning appropriate data to each expert to foster complementary representation learning. Inspired by SimulParallel SGD—which independently trains multiple model replicas on disjoint data subsets and aggregates them via parameter averaging—we view this method as an extreme case of MoE with uniform gating and All-$K$ averaged outputs. Accordingly, in MoE-DisCo, maximizing the distributional divergence among the data subsets assigned to individual experts enhances expert specialization and system diversity, accelerating convergence and improving ensemble effectiveness.  To this end, we employ unsupervised clustering to partition the training corpus: first, semantic embeddings of samples are extracted using a pre-trained model; then, these embeddings are clustered into $K$ groups (where $K$ equals the number of experts) via K-Means, with each cluster assigned to a distinct expert. Clustering naturally groups semantically similar samples, ensuring separation between clusters in the embedding space and satisfying the requirement of ``maximizing distributional divergence.'' This strategy draws inspiration from LRP \cite{2025Latent, gururangan2023scaling} and the theoretical foundations of SimulParallel SGD.

Furthermore, SimulParallel SGD also informs the fusion of the shared backbone: if the sub-datasets are balanced in size, a simple average of backbone parameters suffices to approximate the global optimum; if they are severely imbalanced, a sample-count-weighted average (as in WP-SGD \cite{2020WP}) is required to maintain unbiased gradients.

\begin{mdframed}[leftline=true, rightline=false, topline=false, bottomline=false, linewidth=2.5pt, linecolor=magenta!50, backgroundcolor=gray!10]
In summary, the complete MoE-DisCo training pipeline consists of four key stages: 1. Model Decoupling: The original MoE is decomposed into several  dense submodels, each comprising the full shared backbone plus a single expert; 2. Data Decoupling: Semantic clustering generates divergent sub-datasets, establishing a disentangled alignment between experts and data; 3. Independent Parallel Training: Each submodel is trained independently on its assigned sub-dataset in a fully decentralized manner, with zero communication overhead, and this process can be solved by low-cost hardware; 4. Model Reintegration and Fine-Tune: Experts and the shared backbone are fused, followed by a short global fine-tune phase on the full dataset to restore coordinated gating behavior. The whole process is shown in Figure \ref{fig:training_flow}.
\end{mdframed}

\subsubsection{Construction of Single-Expert Submodels}

Let the MoE model contain $E$ experts. Its parameters split into: (1) \textbf{shared parameters} $\theta_{\text{shared}}$ (embedding, attention, LayerNorm, etc.), and (2) \textbf{expert parameters} $\theta_{\text{exp}} = (\theta_1, \dots, \theta_E)$, where $\theta_k$ is the $k$-th expert’s parameters. Thus, $\Theta = (\theta_{\text{shared}}, \theta_1, \dots, \theta_E)$.

MoE-DisCo constructs lightweight submodels to drastically cut GPU memory during training. Each submodel includes the full $\theta_{\text{shared}}$ and only one expert $\theta_k$. In every MoE layer, the gating mechanism is removed, retaining just a single expert—yielding a compact, dense submodel.

This design sharply reduces model size. For large MoEs, submodels become small enough to train efficiently on low-cost GPUs, shifting workloads from expensive hardware to affordable devices.

Critically, submodels are fully independent during training: no gradient/parameter exchange or synchronization is needed. This eliminates inter-GPU communication overhead and complex multi-GPU coordination, greatly simplifying parallel training. System scalability and deployment flexibility improve significantly. Moreover, since each submodel fits on low-cost hardware, individual device utilization rises, lowering overall training costs.

\subsubsection{Dataset Partitioning}
In MoE-Disco, we propose an unsupervised, clustering-based partitioning method to allocate semantically similar subsets of the training data to individual expert submodels. The approach begins by deriving a fixed-dimensional representation for each input sentence: given a sentence \( x = (x_1, \dots, x_n) \), we encode all its tokens using a pre-trained embedding layer and compute the sentence vector \( h_x \) via mean pooling over the token embeddings,
\begin{equation}
       h_x = \frac{1}{n} \sum_{i=1}^{n} \mathrm{Embedding}(x_i)
\end{equation}
where \( n \) is the sentence length. These sentence vectors are then clustered using K-means with \( K = E \) clusters, where \( E \) denotes the number of experts. 

\subsubsection{Submodel Integrate and Global Training}
After completing independent training of all single-expert submodels, the complete MoE model is constructed and globally optimized through the following steps:
1. Expert layer merging: adopt a "direct integration" strategy to concatenate the trained expert parameters $\theta_1 \sim \theta_E$ into a complete expert layer;
2. Shared parameter fusion: weighted average the shared parameters $\theta_{\text{shared}}^{(k)}$ ($k=1 \sim E$) of all submodels to obtain global shared parameters:
   \begin{equation}
   \theta_{\text{shared}}^* =\sum_{k=1}^E \gamma_k \theta_{\text{shared}}^{(k)}          
   \end{equation}
   $\gamma_k$ is the weight gain from WP-SGD \cite{2020WP}. When subset of dataset size is almost the same, $ \gamma_k = \frac{1}{E}$. The gating  part  is initialized by tensor concatenation.
3. Global fine-tune: assemble the merged parameters ($\theta_{\text{shared}}^*, \theta_{\text{exp}}^*$) into a complete MoE model, and perform joint fine-tune on the original full dataset to alleviate distribution bias between experts and improve overall model performance.

\subsection{MoE-DisCo Algorithm}
Based on above introduction, we give a MoE-DisCo algorithm description, which shown in Algorithm \ref{alg:moe}. 

\begin{algorithm}[t]
\caption{ MoE-DisCo:Mixture-of-Experts with Disentangled Clustering and Coordination }
\label{alg:moe}
\textbf{Require:} Original dataset $\mathcal{D}$; MoE shared parameters $\theta_{\text{shared}}$; Number of experts $E$ and $i$th expert's parameters $\theta_i$; Model $M(\theta,\mathcal{D})$   \\
\textbf{Ensure:} Trained global MoE model $M(\Theta,\mathcal{D)}$

\begin{algorithmic}[1]  
    \State $h_x \leftarrow \text{MeanPool}(\text{Embed}(x))$ for all $x \in \mathcal{D}$ 
    \State $\{\mathcal{D}_1,\dots,\mathcal{D}_E\} \leftarrow \text{K-means}(\{h_x\}, K=E)$ 
    
    \For{$k \leftarrow 1$ to $E$}
        \State $\theta_{\text{shared}}^{(k)} \leftarrow \theta_{\text{shared}}$ 
        \State $\Theta_k \leftarrow (\theta_{\text{shared}}^{(k)}, \theta_k)$ 
        \State\text{Train} $ M(\Theta_k, \mathcal{D}_k)$ 
    \EndFor
    
    \State $\theta_{\text{exp}}^* \leftarrow \text{Concat}(\theta_1,\dots,\theta_E)$ 
    \State $\theta_{\text{shared}}^* \leftarrow \frac{1}{E}\sum_{k=1}^E \theta_{\text{shared}}^{(k)}$
    \State $\Theta \leftarrow (\theta_{\text{shared}}^*, \theta_{\text{exp}}^*)$ 
    \State \text{Fine-tune} $M(\Theta, \mathcal{D})$ 
    
    \Return $M(\Theta, \mathcal{D})$ 
\end{algorithmic}
\end{algorithm}
 
 \begin{figure*}[hbpt!]
  \centering
  \subfloat[]{
    \includegraphics[width=0.27\linewidth]{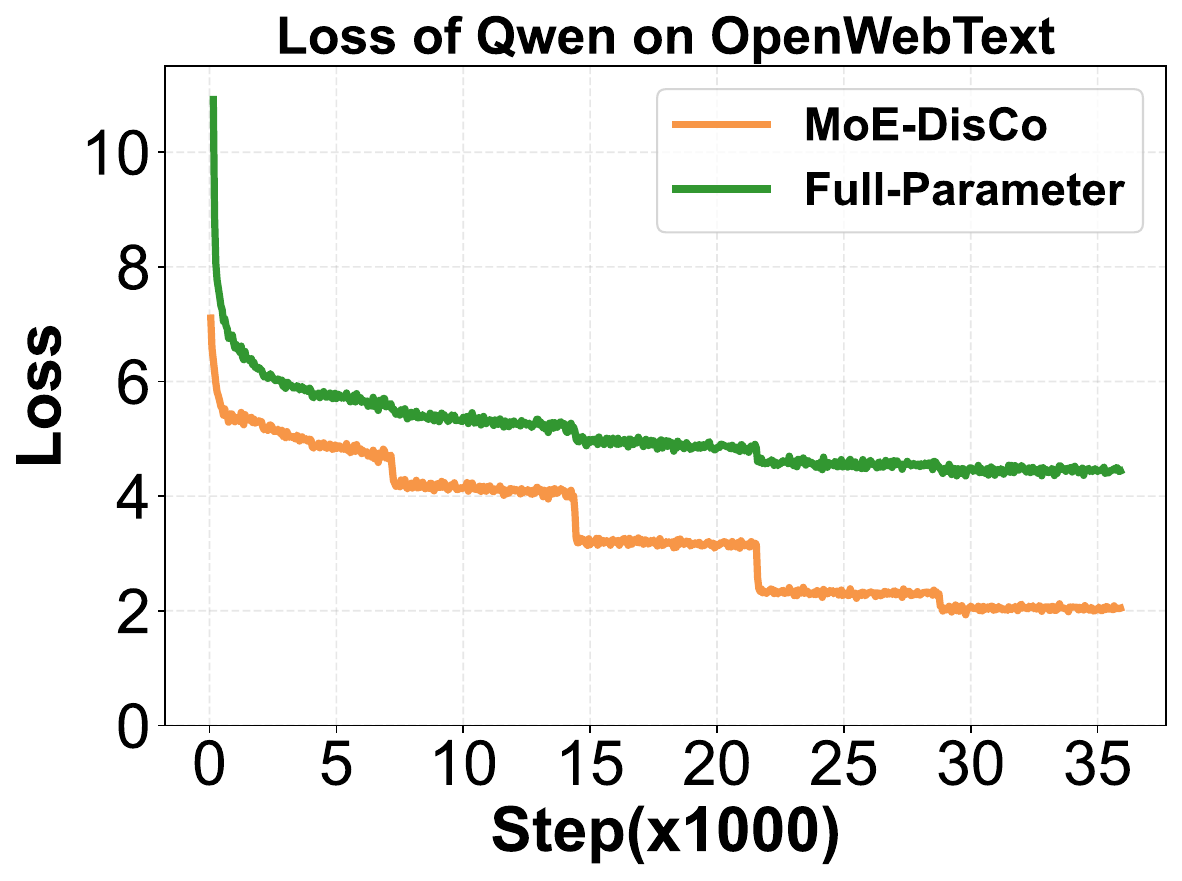}
  }\hfill
  \subfloat[]{
    \includegraphics[width=0.27\linewidth]{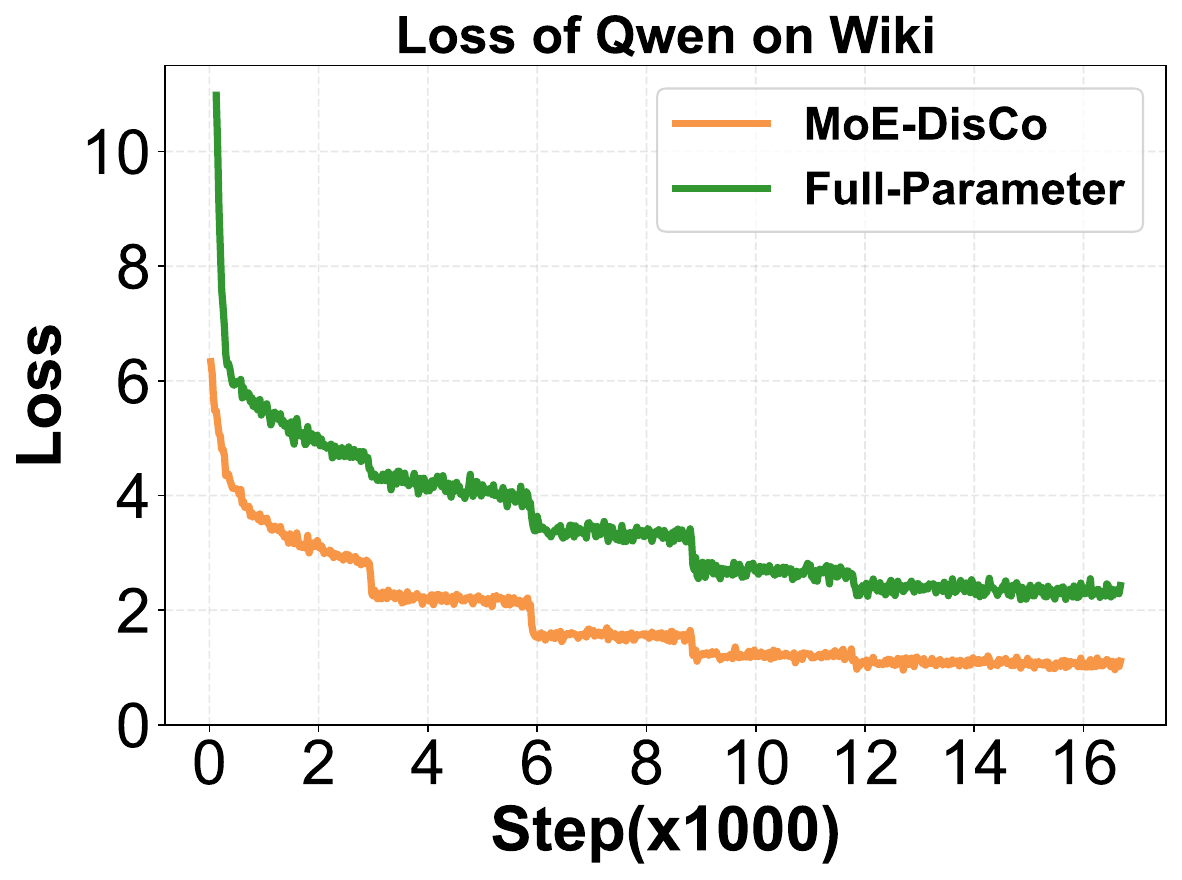}
  }\hfill
  \subfloat[]{
    \includegraphics[width=0.27\linewidth]{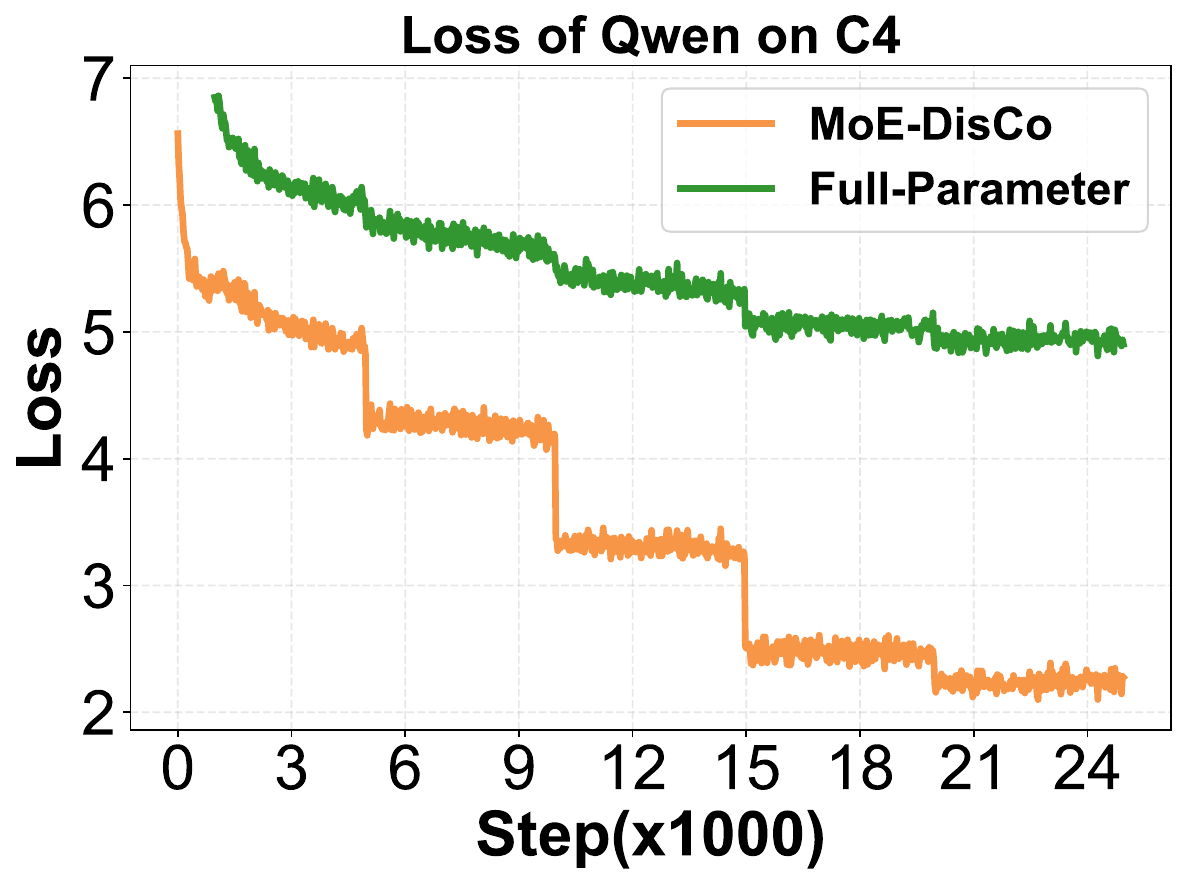}
  }\hfill
  \subfloat[]{
    \includegraphics[width=0.27\linewidth]{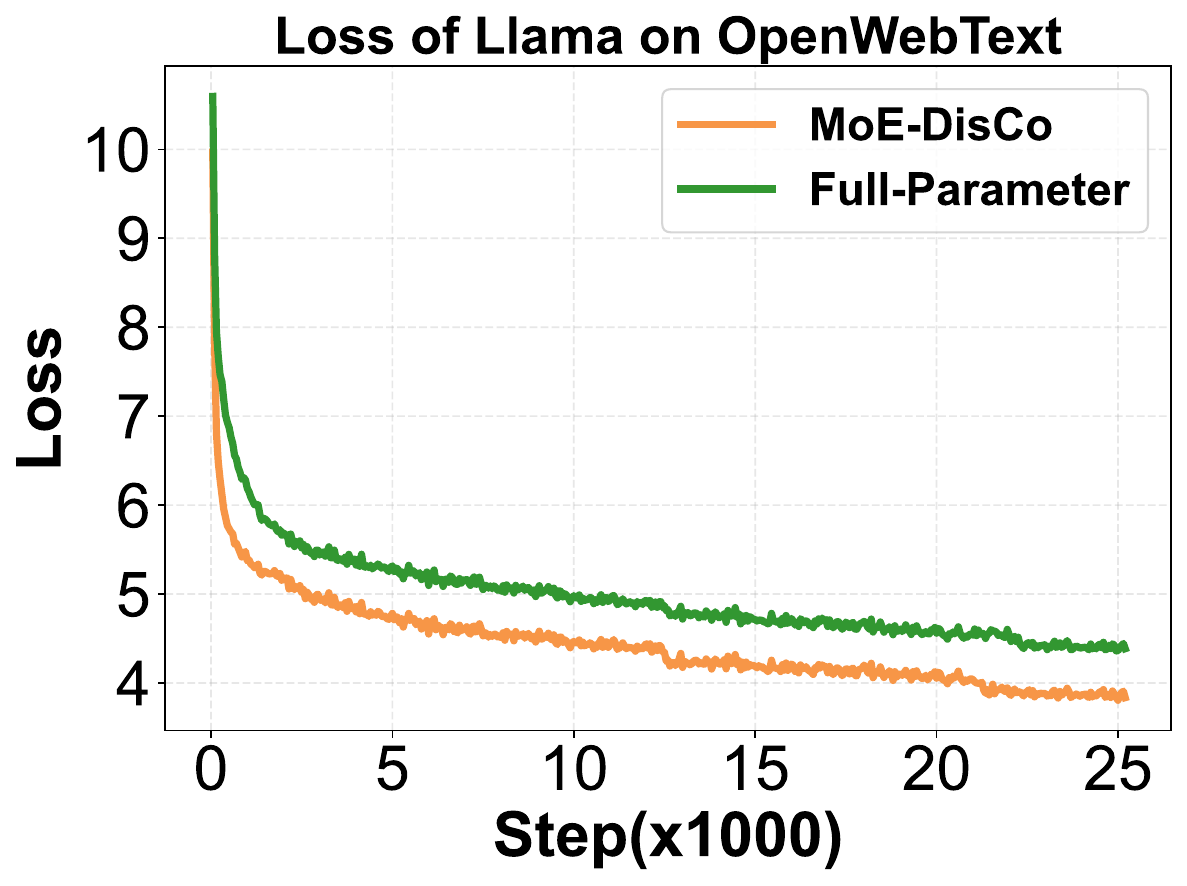}
  }\hfill
  \subfloat[]{
    \includegraphics[width=0.27\linewidth]{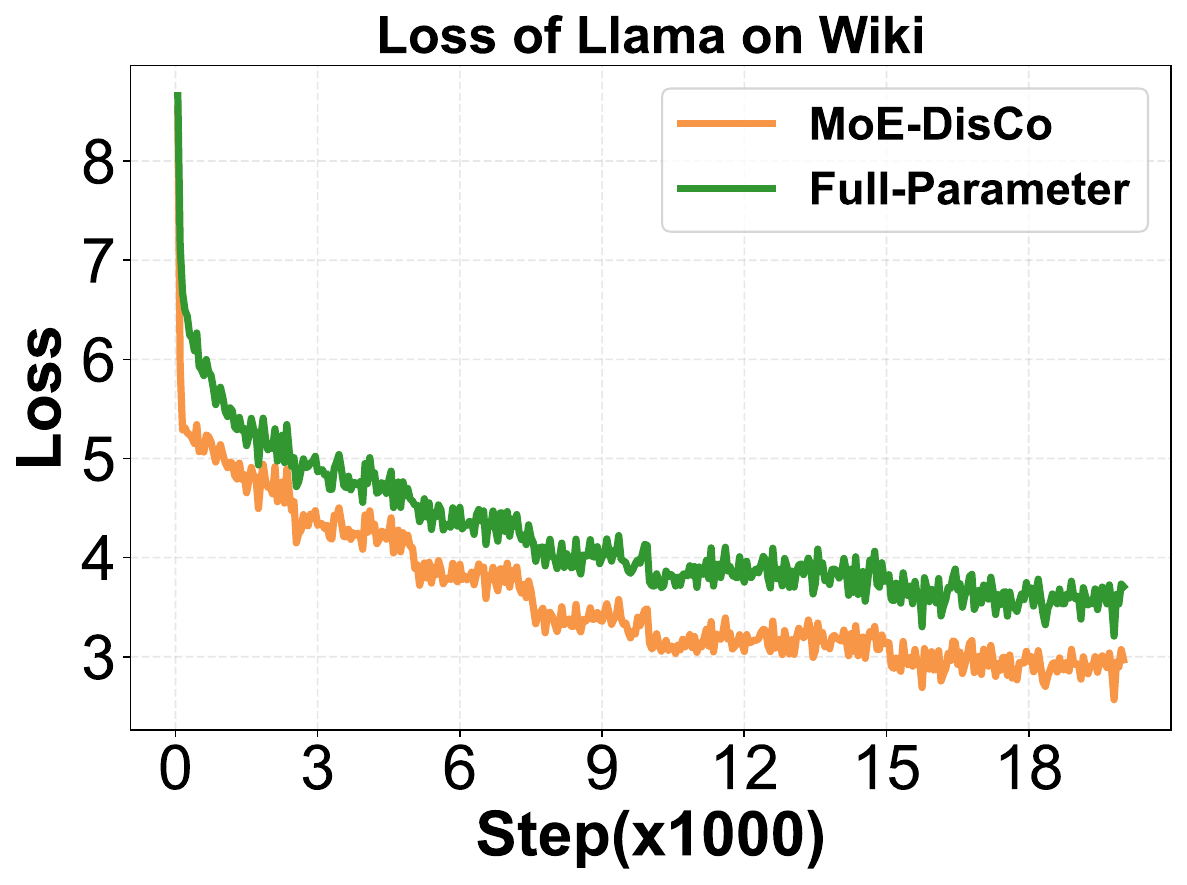}
  }\hfill
  \subfloat[]{
    \includegraphics[width=0.27\linewidth]{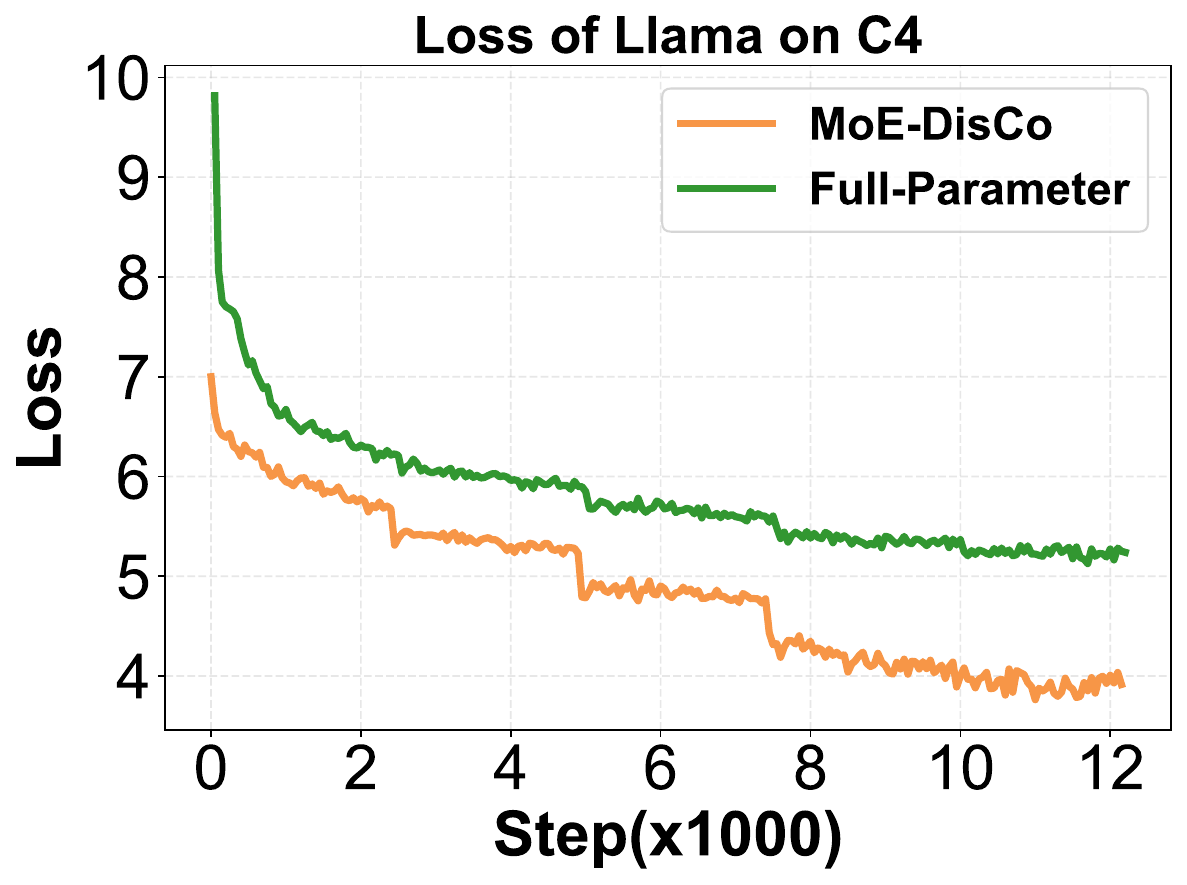}
  }\hfill

  \caption{ Loss  trends between MoE-Disco on fine-tune stage and Full-Parameter across different datasets}
  \label{fig:loss_trends}
\end{figure*}

\section{Experiments}
In this work, we conduct a systematic comparison between MoE-DisCo and a full-parameter trained MoE baseline, evaluating both model performance and training cost.
For performance evaluation, we compare the resulting models in terms of language modeling quality (measured by training loss, PPL and downstream task) demonstrating that MoE-DisCo preserves—or even enhances—algorithmic effectiveness relative to standard MoE training.
For economy cost analysis, we focus specifically on resource consumption during the fine-tune phase, as this is the only stage requiring high-cost GPUs (e.g., A100). The main training phase of MoE-DisCo, by contrast, runs entirely on low-cost hardware, making fine-tune the primary determinant of its overall infrastructure expense. For ablation study, we will show the influence of k-means cluster and the number of experts.

\subsection{Experimental Setup}
\label{sec:baselines_models}

\subsubsection{Models}
\label{sec:models}

\textbf{Qwen1.5-MoE-2.7B.}
The Qwen1.5-MoE-2.7B model activates approximately 2.7 billion parameters during inference, while achieving performance comparable to dense models with around 7 billion parameters, such as Mistral-7B. In our experiments, we set the number of experts to 4 while keeping all other configurations identical to the original model. We use abbr. Qwen in following parts.

\textbf{LLaMA-MoE-3.5B.}
The LLaMA-MoE-3.5B model is built upon the LLaMA architecture and incorporates a Mixture-of-Experts design to improve parameter efficiency and scalability. In our experiments, we set the number of experts to 4 while keeping all other configurations identical to the original model. We use abbr. Llama in following parts.

\subsubsection{Datasets}
Experiments use three public standard datasets, as follows: 1. C4: A large-scale English corpus with strict filtering and cleaning; 2. WikiText-2: A high-quality English dataset compiled from Wikipedia, commonly used for language model benchmark evaluation; 3. OpenWebText: An open web text collection constructed following the GPT-2 training data.

\subsubsection{ Evaluation Metrics}

We evaluate MoE-DisCo against the full-parameter trained MoE baseline using three key criteria: (1) language modeling capability, measured by training loss , PPL and downstream tasks; (2) training efficiency, assessed via the number of training steps required to reach a target loss and the total usage time of high-cost GPUs. (3) The economy cost of training a MoE model in dollar.

\subsection{Performance Comparison}
 Table \ref{tab:performance_comparison_combined} presents a comparison of training efficiency and convergence performance between MoE-DisCo and full-parameter training across two MoE architectures (Qwen and LLaMA) and three datasets (C4, WikiText-2, and OpenWebText). The results show that MoE-DisCo substantially reduces the number of training steps required to reach the same training loss. For instance, on the Qwen architecture, the steps to achieve convergence decrease from 21150, 12500, and 28600 to 4100, 3150, and 6650 on the C4, WikiText-2, and OpenWebText datasets, respectively, representing more than a fourfold reduction. A similar trend is observed for the LLaMA architecture. Moreover, at the same training loss, the PPL of models trained by MoE-DisCo metgod is comparable to, and in some cases lower than that of  models trained by full-parameter method, indicating that the method accelerates training without compromising language modeling performance.

\begin{figure*}[hbpt!]
  \centering
  \subfloat[]{
    \includegraphics[width=0.3\linewidth]{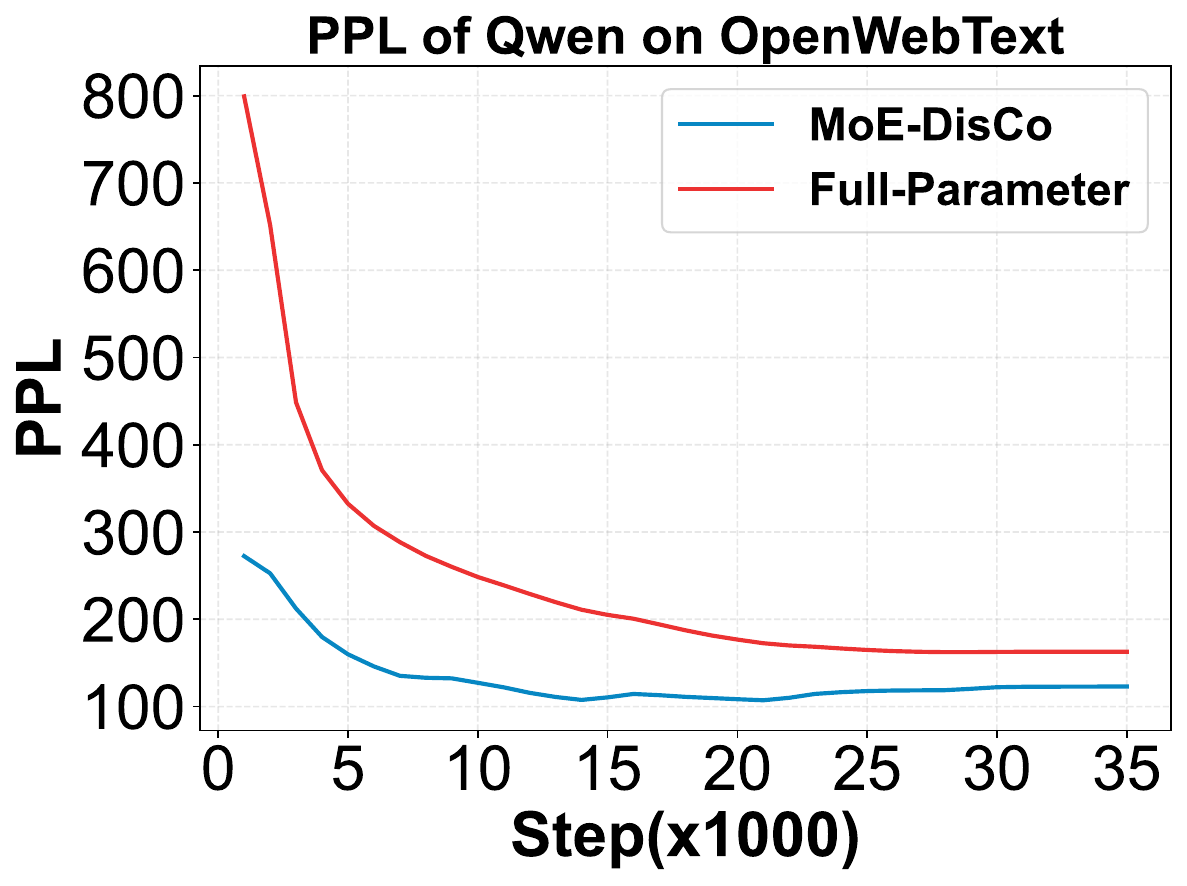}
  }\hfill
  \subfloat[]{
    \includegraphics[width=0.3\linewidth]{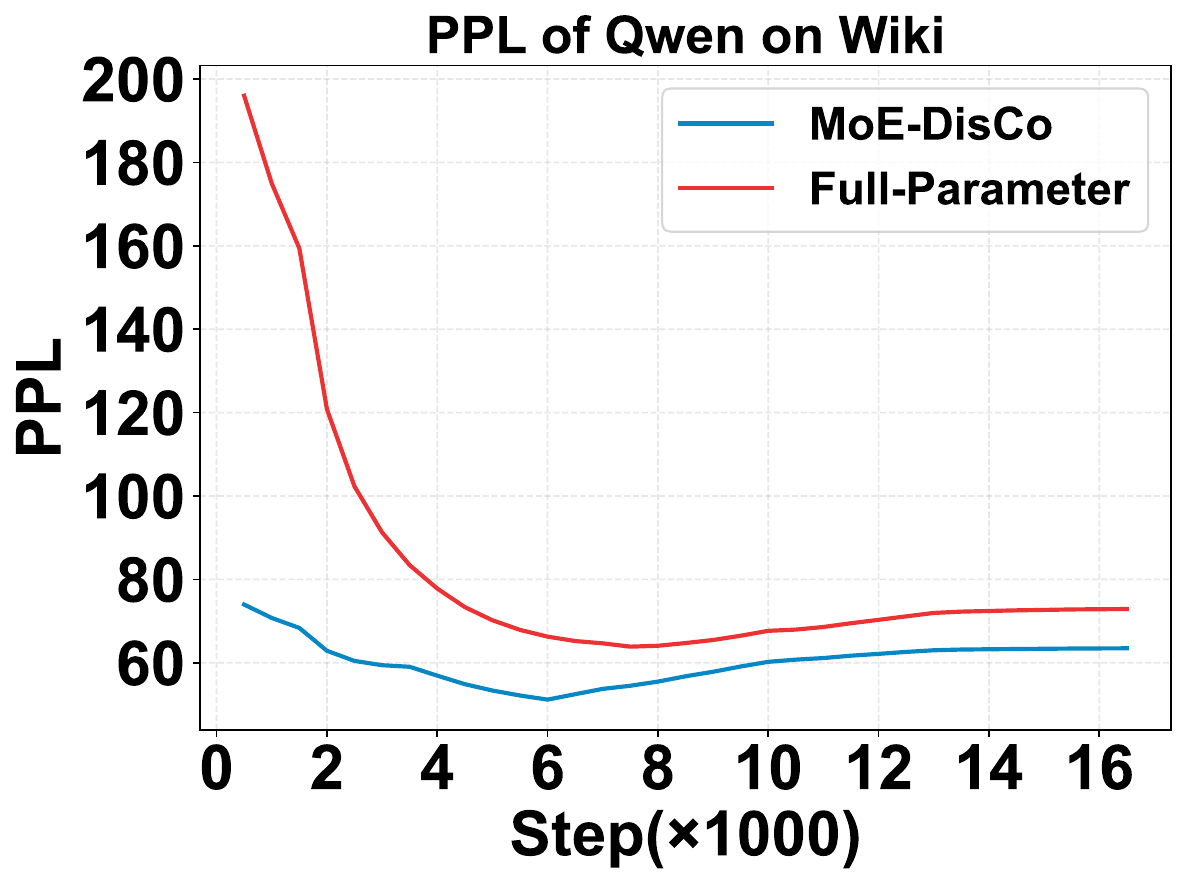}
  }\hfill
  \subfloat[]{
    \includegraphics[width=0.3\linewidth]{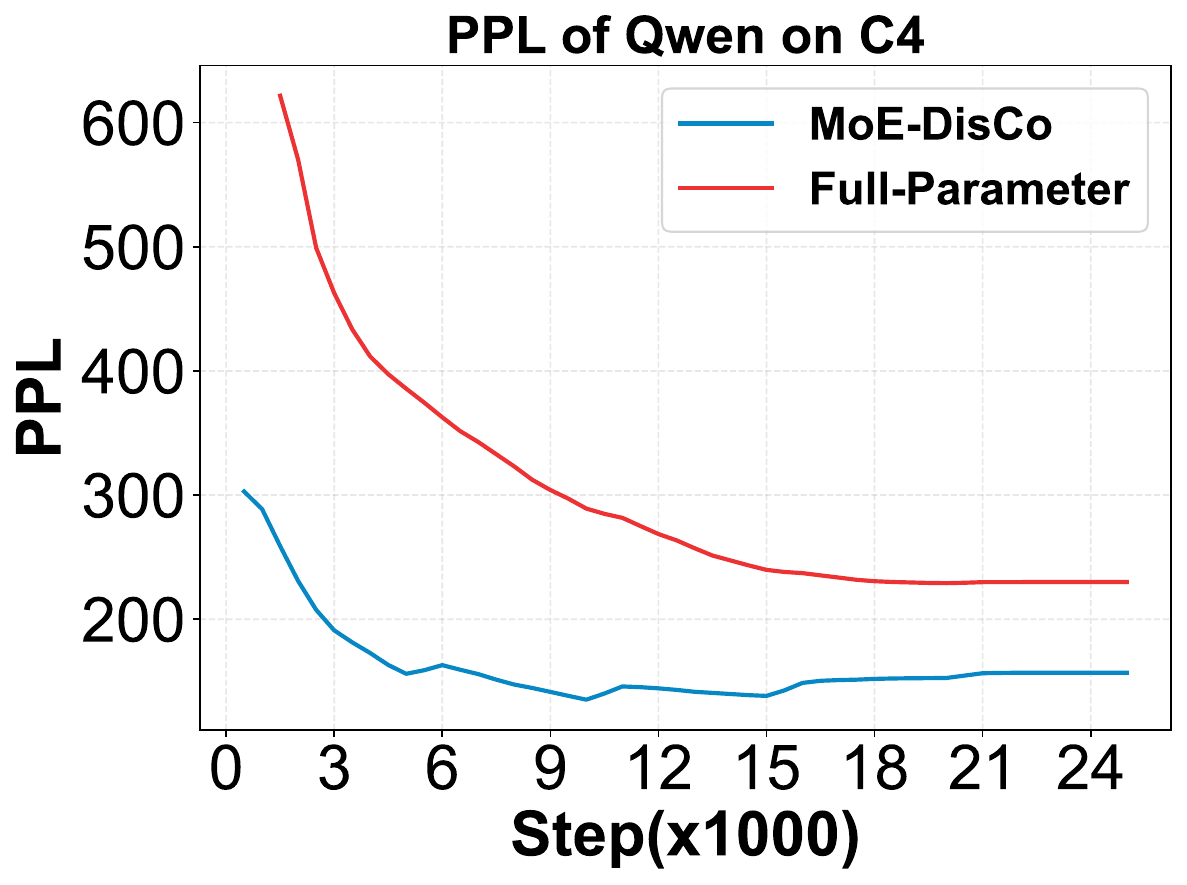}
  }\hfill
  \subfloat[]{
    \includegraphics[width=0.3\linewidth]{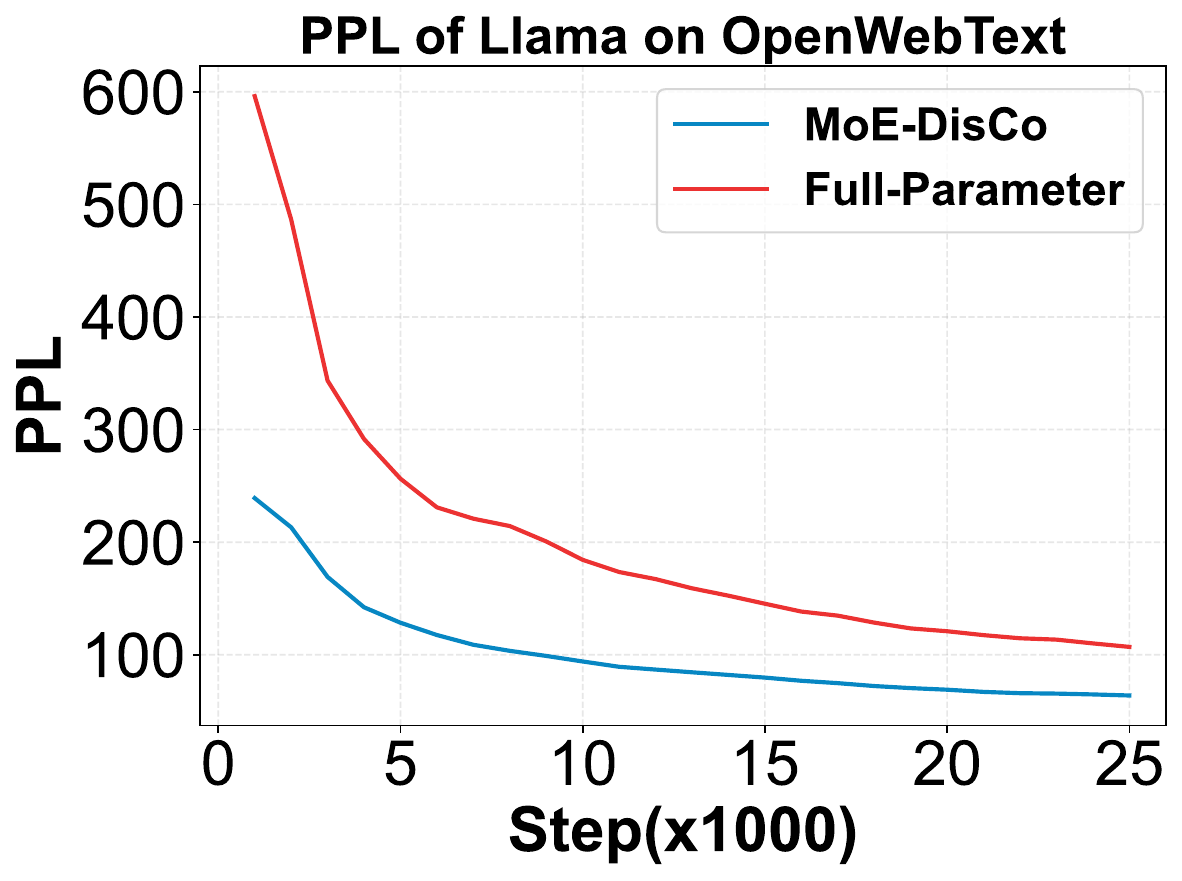}
  }\hfill
  \subfloat[]{
    \includegraphics[width=0.3\linewidth]{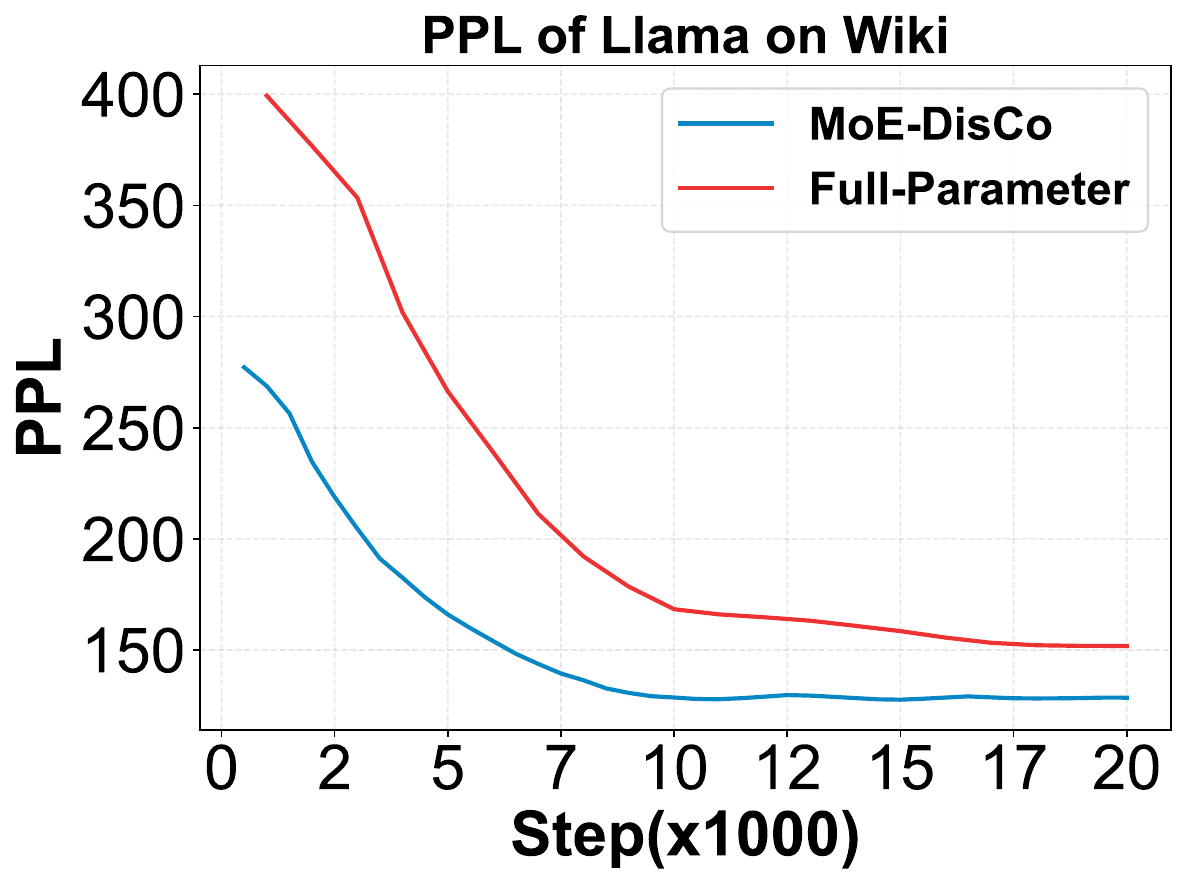}
  }\hfill
  \subfloat[]{
    \includegraphics[width=0.3\linewidth]{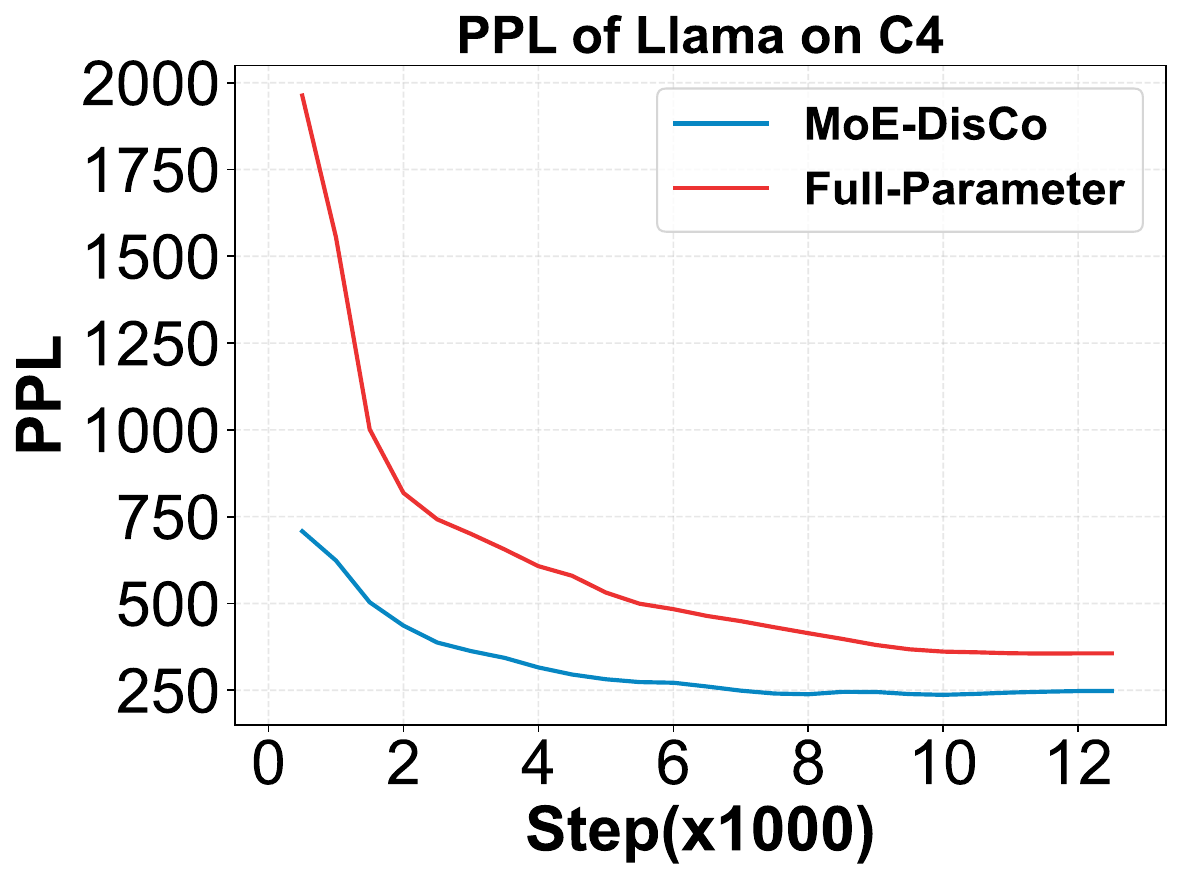}
  }\hfill
  
  \caption{ PPL  trends between MoE-Disco on fine-tune stage and Full-Parameter across different datasets}
  \label{fig:ppl_curves}
\end{figure*}

Figures \ref{fig:loss_trends} and \ref{fig:ppl_curves} show the training loss and ppl curves of the Full-Parameter and MoE-DisCo  fine-tune stage methods on three datasets, respectively. We use fine-tune stage of MoE-DisCo because this part is the most expensive stage (Cost information is shown in Economic Cost Analysis section). It can be observed that, given the same training duration, models trained with the MoE-DisCo method achieve significantly better performance than those trained with full-parameter training. If the target is to reach the same model performance, the MoE-DisCo method substantially reduces the required training time. This indicates that the proposed method can markedly lower training economy costs without sacrificing model performance.   The whole training process including submodel training is shown in Cost Analysis part and Appendix.

\begin{table}[t!]
  \centering
  \small
  \setlength{\tabcolsep}{3pt}
  \renewcommand{\arraystretch}{1.8}
  \scalebox{0.8}{
  \begin{tabular}{lccccccc}
    \toprule
    \multirow{2}{*}{\textbf{Model}} & \multirow{2}{*}{\textbf{Data}} & \multicolumn{3}{c}{\textbf{Full-Param}} & \multicolumn{3}{c}{\textbf{MoE-DisCo}} \\
    \cmidrule(lr){3-5} \cmidrule(lr){6-8}
    & & Step & Loss & PPL & Step & Loss & PPL \\
    \midrule
    Qwen & C4 & 21,150 & 4.954 & 230.32 & \cellcolor{blue!15}4,100 &  4.925 &  165.86 \\
         & WikiText-2  & 12,500 & 2.303 & 71.89  & \cellcolor{blue!15}3,150 &  2.2964 &  57.55 \\
         & OpenWebText          & 28,600 & 4.606 & 162.51 &  \cellcolor{blue!15}6,650 &  4.588 &  134.31 \\
    \midrule
    Llama& C4          & 11,750 &  5.274 & 356.28  & \cellcolor{blue!15}4,150 & 5.276 &  296.35 \\ 
         & WikiText-2  & 15,000 & 3.805 &  160.91 & \cellcolor{blue!15}5,300 & 3.794 & 163.27 \\ 
         & OpenWebText & 21,000 &  4.536 & 114.93  &  \cellcolor{blue!15}7,800 & 4.541 & 106.82 \\ 
    \bottomrule
  \end{tabular}
  }
  \caption{Performance comparison between MoE-DisCo and full-parameter training across two MoE architectures. The  table show that the number of steps required for MoE-DisCo to achieve full-parameter convergence result.}
  \label{tab:performance_comparison_combined}
\end{table}

\subsection{Downstream Task Performance}

Table~\ref{tab:downstream_performance} reports the downstream task performance of MoE-DisCo and full-parameter training. All models are pretrained on OpenWebText and evaluated on multiple downstream tasks. ARC-e and MMLU are evaluated in the 5-shot setting, while HellaSwag and PIQA are evaluated in the zero-shot setting.

As shown in the table, MoE-DisCo consistently achieves better performance than full-parameter training on both few-shot (ARC-e, MMLU) and zero-shot (HellaSwag, PIQA) downstream tasks.
Overall, the results show that MoE-DisCo achieves better downstream task performance than full-parameter training across the evaluated benchmarks.


\begin{table}[htbp]  
  \centering
  \scalebox{.8}{ 
  \small  
  \begin{tabular}{@{}lcccc@{}} 
    \toprule
    \textbf{Method} & \textbf{ARC-e (5)} & \textbf{MMLU (\%)} & \textbf{HellaSwag (\%)} & \textbf{PIQA (\%)} \\
    \midrule
    Full-Param  & 27.9 & 23.0 & 27.45 & 52.25 \\
    MoE-DisCo   & \cellcolor{blue!15}29.1 & \cellcolor{blue!15}25.3 & \cellcolor{blue!15}29.45 & \cellcolor{blue!15}57.03 \\
    \bottomrule
  \end{tabular}
  }
  \caption{Downstream task performance of Qwen1.5-MoE-2.7B on ARC-E, MMLU, Hellaswag and PIQA.
(ARC-r and MMLU are evaluated under 5-shot settings; Hellaswag and PIQA under zero-shot.).}
  \label{tab:downstream_performance}
\end{table}
\subsection{Economic Cost Analysis}
Training large-scale models, especially those based on MoE architectures, incurs substantial computational and memory costs. 
This section analyzes the economic advantages of the proposed MoE-DisCo training strategy and compares it with conventional full-parameter training methods.

Following the market pricing of major cloud service providers and considering the hardware environment used in our experiments, we adopt the following GPU rental prices as the cost estimation baseline:RTX 4090: \$0.35/ GPU$\cdot$hour;  A100 (80GB): \$2.28 / GPU$\cdot$hour.

\begin{table*}[t!]
  \centering
  \small
  \setlength{\tabcolsep}{3pt} 
  \renewcommand{\arraystretch}{1.8} 
  \scalebox{0.8}{
  \begin{tabular}{lcccccccccccc}
    \toprule
    \multirow{2}{*}{\textbf{Model}} & \multirow{2}{*}{\textbf{Dataset}} & \multicolumn{3}{c}{\textbf{Full-Parameter}} & \multicolumn{8}{c}{\textbf{MoE-DisCo}} \\
    \cmidrule(lr){3-5} \cmidrule(lr){6-13}
    & & Cost(\$) & Time(h) & Platform & S-Cost(\$) & S-Time(h) & S-Platform & F-Cost(\$) & F-Time(h) & F-Platform & Total Cost(\$) & Total Time(h) \\
    \midrule
    \multirow{3}{*}{Qwen} 
      & C4          & 22.5036 & 9.87  & A100*1 
                    & 2.9260 & 2.09  & \cellcolor{blue!15}RTX 4090*4 
                    & 3.9444 & 1.73 & A100*1 
                    & \cellcolor{blue!15}6.8704 & \cellcolor{blue!15}3.82 \\
      & Wiki        & 6.9312  & 3.04   & A100*1 
                    & 1.7920  & 1.28   & \cellcolor{blue!15}RTX4090*4 
                    & 1.5504  & 0.68  & A100*1 
                    & \cellcolor{blue!15}3.3424 & \cellcolor{blue!15}1.96 \\
      & OpenWebText & 29.9136 & 13.12  & A100*1 
                    & 4.3680  & 3.12   & \cellcolor{blue!15}RTX4090*4 
                    & 6.5436  & 2.87   & A100*1 
                    & \cellcolor{blue!15}10.9116 & \cellcolor{blue!15}5.99 \\
    \midrule
    \multirow{3}{*}{Llama} 
      & C4          & 12.8592 & 5.64  & A100*1 
                    & 2.1560  & 1.54  & \cellcolor{blue!15}RTX4090*4 
                    & 4.5828 & 2.01 & A100*1 
                    & \cellcolor{blue!15}6.7388 & \cellcolor{blue!15}3.55 \\
      & Wiki        & 16.9860 & 7.45  & A100*1 
                    & 2.0440 & 1.46  & \cellcolor{blue!15}RTX4090*4 
                    & 6.0876 & 2.67 & A100*1 
                    & \cellcolor{blue!15}8.1316 & \cellcolor{blue!15}4.13 \\
      & OpenWebText & 32.1252 & 14.09 & A100*1 
                    & 4.2140 & 3.01  & \cellcolor{blue!15}RTX4090*4 
                    & 10.9212 & 4.79 & A100*1 
                    & \cellcolor{blue!15}15.1352 & \cellcolor{blue!15}7.80 \\
    \bottomrule
  \end{tabular}
  }
  \caption{Comparison of Full-Parameter and MoE-DisCo training costs, times, and platforms for Qwen and Llama across three datasets. In MoE-DisCo, Fine-tune Cost, Fine-tune Time, and Platform are shown separately. S- means submodel training. F- means fine-tune.  The S-Cost is the all cost in submodel training.}
  \label{tab:moe_costs_combined_platform}
\end{table*}


Based on the results in Table~\ref{tab:moe_costs_combined_platform}, MoE-DisCo achieves model performance comparable to—sometimes slightly better than—full-parameter training, while substantially reducing both cost and training time. The approach decouples MoE training into two phases: (1) Submodel Training (S-phase) and (2) Lightweight Fine-tuning (F-phase).

During the S-phase, all submodels are trained independently and in parallel on a low-cost RTX 4090 platform. To simplify accounting and provide a conservative upper bound on resource usage, the total S-phase cost and time are computed as four times the longest individual submodel training duration, reflecting worst-case synchronization overhead. Due to the low per-unit cost of RTX 4090 and the reduced computational footprint of submodels, this phase incurs only \$1.79–\$4.37 across all experiments. Critically, because training is fully parallelized, marginal costs scale sublinearly—effectively near-constant—with the number of experts.

The F-phase requires only a brief synchronized fine-tune step on a single A100 GPU, sufficient to recover (or slightly exceed) the convergence of full-parameter training. This phase costs \$1.55–\$10.92.

Overall, MoE-DisCo achieves significant savings across all settings. On Qwen, total costs on C4, WikiText-2, and OpenWebText are \$6.87, \$3.34, and \$10.91—69.5\%, 51.8\%, and 63.5\% lower than full-parameter baselines (\$22.50, \$6.93, \$29.91)—with training time reduced from 9.87h, 3.04h and 13.12h to 3.82h, 1.96h, and 5.99h. On Llama, MoE-DisCo costs \$6.74, \$8.13, and \$15.14—47.6\%, 52.2\%, and 52.9\% less than full-parameter training and reduces training time from 5.64h, 7.45h, and 14.09h to 3.55h, 4.13h, and 7.80h.

Despite drastic cost reduction, final performance remains on par with or marginally better than full-parameter training. By shifting the majority of computation from expensive A100 platform to low-cost RTX 4090 platform via a “decoupled training + lightweight reintegration” strategy, MoE-DisCo offers a practical, reproducible, and highly cost-efficient framework for training large MoE
models—particularly valuable for academic and industrial teams under resource constraints.

\begin{figure}[t]
  \centering
  \includegraphics[width=0.35\textwidth]{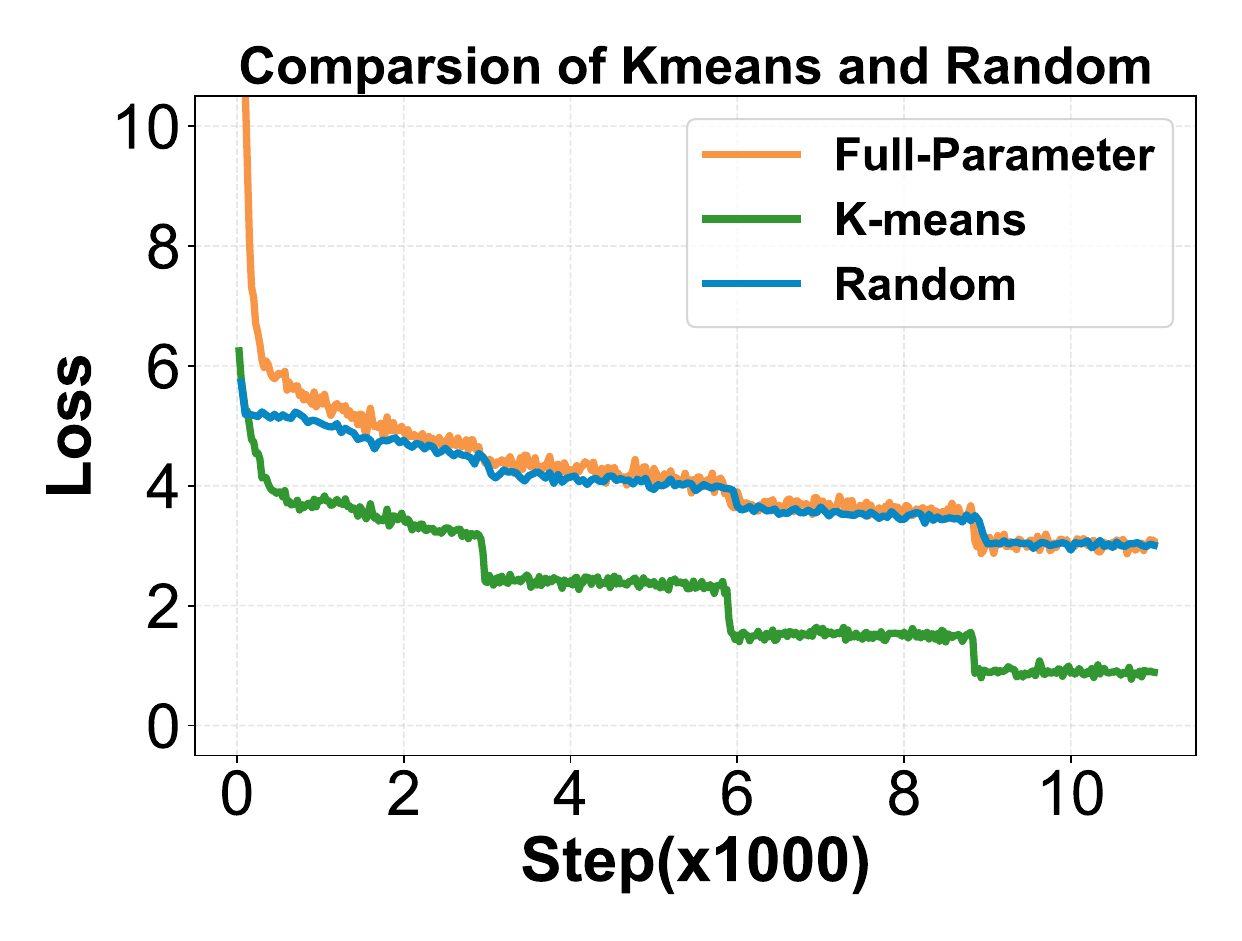} 
   
  \caption{In the Qwen model on C4, replacing the cluster operation in MoE-DisCo with random data assignment causes the MoE training performance during the fine-tune stage to degrade to that of full-parameter training.}
  \label{fig:ablation  cluster}
\end{figure}
\begin{figure}[t]
  \centering
  \includegraphics[width=0.35\textwidth]{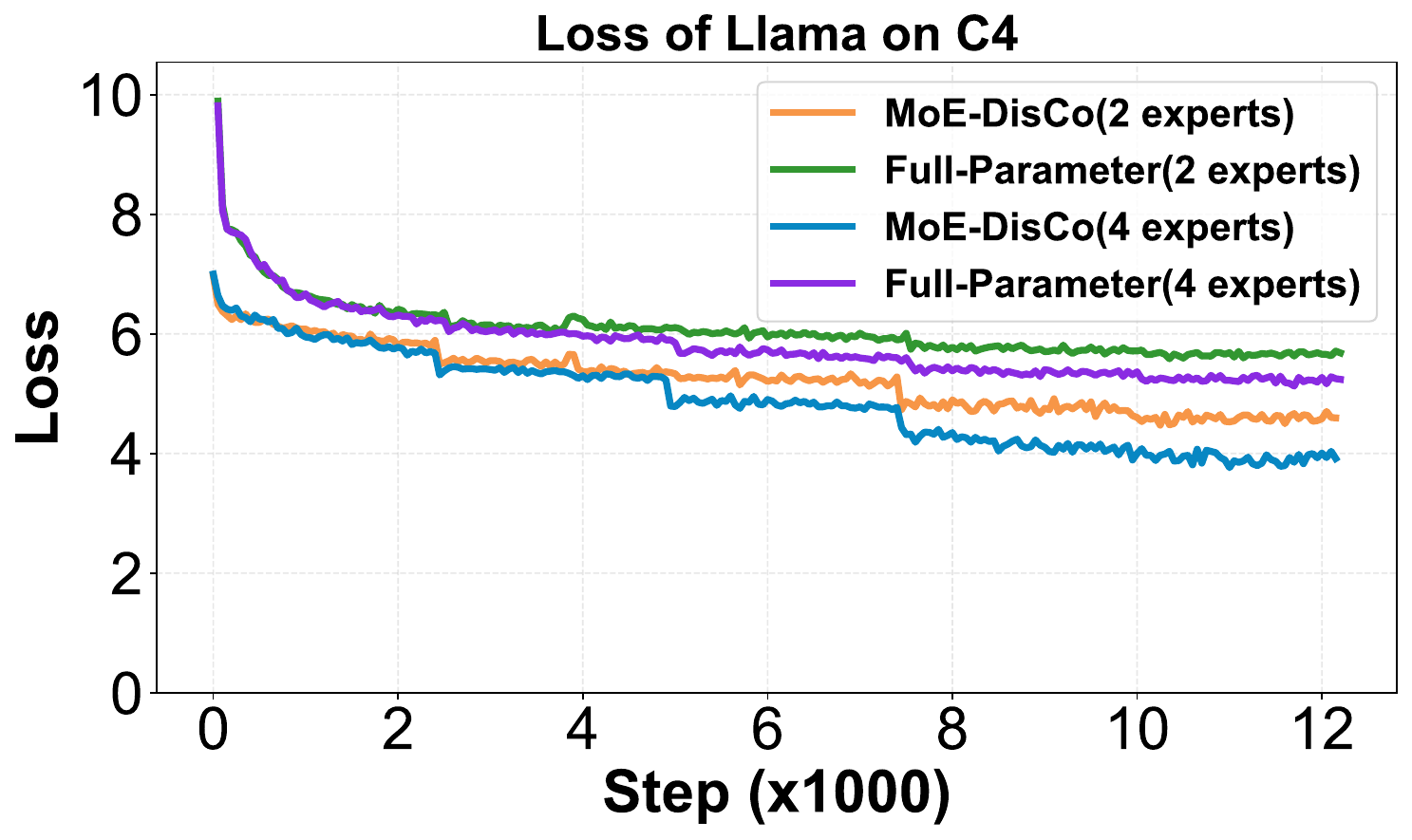} 

  \caption{The MoE-DisCo performance on Different MoE model with 2 and 4 experts}
  \label{fig:ablation  experts}
\end{figure}

\subsection{Ablation Study}
\subsubsection{Cluster in MoE-DisCo}
In this ablation study, we validate the importance of the clustering operation in the MoE-DisCo framework. Specifically, during the training of the Qwen model on C4, we removed the original k-means clustering step and replaced it with a random assignment of data to different experts for submodel training with the same amount of dataset. The results, Figure \ref{fig:ablation  cluster},show that substituting k-means clustering with random assignment significantly degrades training efficiency, and the final performance essentially regresses to that of full-parameter training. We therefore conclude that k-means clustering is an important component of MoE-DisCo.

\subsubsection{Multi-experts in MoE-DisCo}
In this ablation study, we evaluate MoE-DisCo’s performance with varying expert counts to assess how architectural capacity affects training efficiency and model quality of MoE-DisCo. Specifically, we compare the convergence of the Llama model on C4 using 2 and 4 experts. As shown in Figure \ref{fig:ablation experts}, MoE-DisCo achieves faster convergence and better final performance in both cases. Notably, the 4-expert model yields slightly lower loss and perplexity than the 2-expert version. Results suggest that MoE-DisCo achieves high training efficiency and strong performance across different numbers of experts.

\section{Conclusion}
This study addresses the challenges of high cost of MoE training  by proposing a method that applies  BCD and SimulParallel SGD to MoE training, i.e., MoE-DisCo. Compared with traditional full-parameter training, this study verifies the feasibility and advantages of the MoE-DisCo method in reducing  economy cost, decreasing reliance on high-cost GPUs. It achieves large-scale model training under low computational resource conditions, providing theoretical and practical guidance for large model training in resource-constrained scenarios.
\section{Limitation}
- Limitations: Due to resource constraints, the effectiveness of the method has not been verified on larger-scale models (e.g., 10B+ parameters);
- Future work: Expand model scale and dataset scope, introduce more evaluation metrics; explore dynamic expert number adjustment strategies to further improve training efficiency and model performance.

\appendix

\section{ More of Dataset Partitioning}
\begin{figure}[h]
  \centering
  \includegraphics[width=0.7\columnwidth]{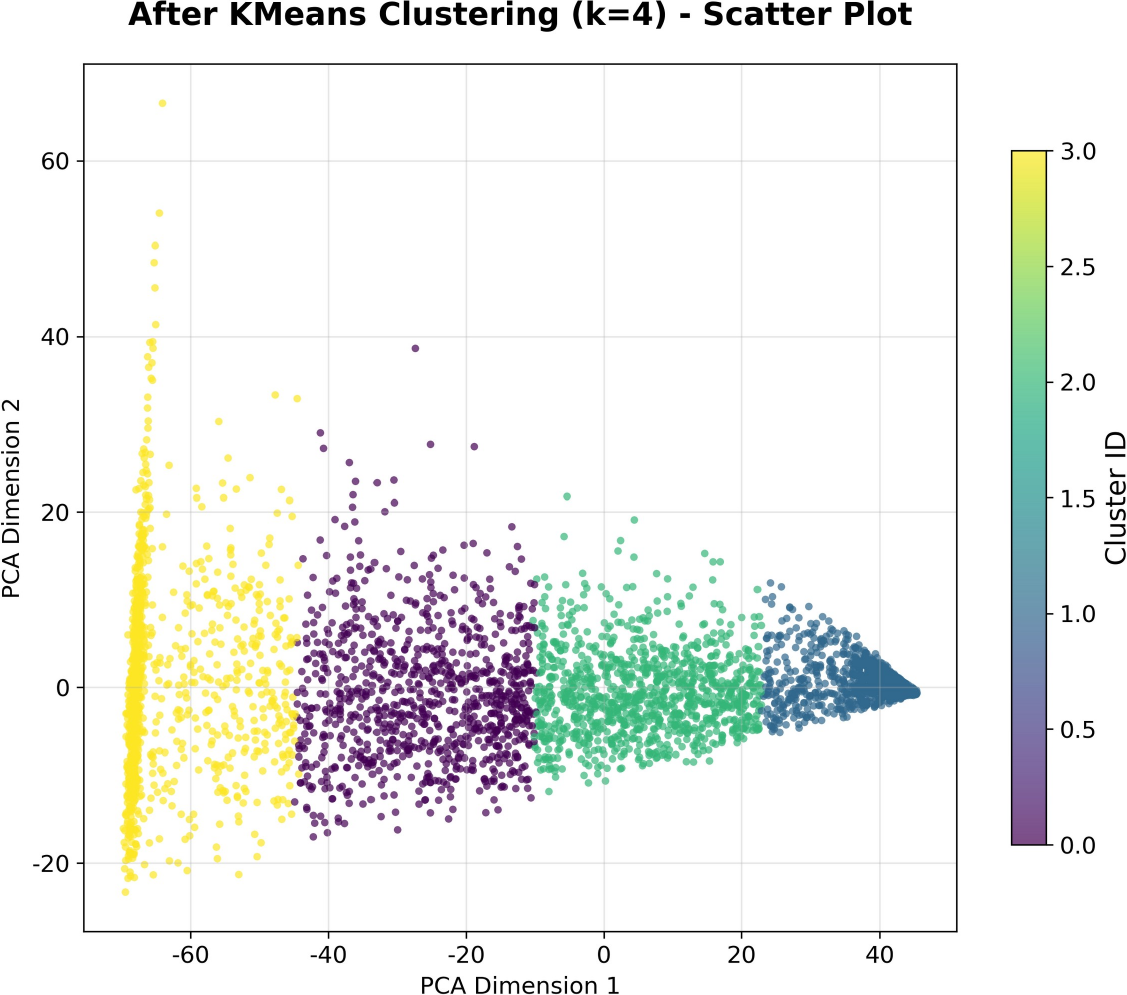}  
  \caption{Scatter plot distribution of the C4 dataset on four clusters after clustering (PCA dimensionality reduction to 2D).}
  \label{fig:cluster_scatter}
\end{figure}
In MoE-Disco, we propose an unsupervised, clustering-based partitioning method to allocate semantically similar subsets of the training data to individual expert submodels. The approach begins by deriving a fixed-dimensional representation for each input sentence: given a sentence \( x = (x_1, \dots, x_n) \), we encode all its tokens using a pre-trained embedding layer and compute the sentence vector \( h_x \) via mean pooling over the token embeddings,
\begin{equation}
\begin{small}
       h_x = \frac{1}{n} \sum_{i=1}^{n} \mathrm{Embedding}(x_i)
\end{small}
\end{equation}
where \( n \) is the sentence length. These sentence vectors are then clustered using K-means with \( K = E \) clusters, where \( E \) denotes the number of experts.   
The clustering objective minimizes the within-cluster sum of squared distances to the respective centroids:
\begin{equation}
    \min_{\{C_k\}_{k=1}^K} \sum_{k=1}^K \sum_{h_x \in C_k} \| h_x - \mu_k \|^2,
\end{equation}
with \( C_k \) representing the set of sentence vectors assigned to cluster \( k \) and \( \mu_k \) its centroid. Finally, the resulting clusters are used to partition the original dataset into \( E \) semantically coherent sub-datasets, each of which is assigned to train a distinct single-expert submodel. 
We show a example of clustering dataset in Figure \ref{fig:cluster_scatter}.

\section{Code Link}
https://anonymous.4open.science/r/MoE-DisCo-4835/
\section{Hyperparameter Settings}
\label{app:hyperparams}

All experiments in this work are conducted with the following hyperparameters :

\begin{table}[htbp]
\centering
\small
\scalebox{1.2}{ 
\begin{tabular}{@{}ll@{}}
\toprule
\textbf{Hyperparameter} & \textbf{Value} \\
\midrule
Optimizer               & AdamW \\
Learning rate           & $1 \times 10^{-4}$ \\
Learning rate scheduler & constant \\
Batch size              & 16 \\
Data Type               & \texttt{bfloat16} \\   
Sequence length         & 1024 \\
\bottomrule
\end{tabular}
}
\caption{Hyperparameter settings used in submodel training.}
\label{tab:hyperparameters}
\end{table}
\begin{table}[htbp]
\centering
\small
\scalebox{1.2}{ 
\begin{tabular}{@{}ll@{}}
\toprule
\textbf{Hyperparameter} & \textbf{Value} \\
\midrule
Optimizer               & AdamW \\
Learning rate           & $3 \times 10^{-4}$ \\
Weight decay            & 0.01 \\
Warmup ratio            & 0.03 \\
Learning rate scheduler & Cosine \\
Batch size              & 16 \\
Data Type               & \texttt{bfloat16} \\   
Sequence length         & 1024 \\
\bottomrule
\end{tabular}
}
\caption{Hyperparameter settings used in full-parameter training and MoE-DisCo fine-tune.}
\label{tab:hyperparameters}
\end{table}

\section{Submodel Training Details}

In MoE-DisCo, each submodel is trained on a low-cost GPU.
For the $k$-th submodel, we retain the full shared backbone and keep only the $k$-th expert in every MoE layer. This results in a standard dense model with the same depth and hidden dimensionality as the original MoE.
Although sub-expert training requires a non-trivial amount of time, it incurs relatively low monetary cost, as each submodel can be trained independently on affordable hardware and fully in parallel.
A detailed breakdown of training steps and cost is provided in Table~\ref{tab:subexpert_cost}.
\begin{table}[t]
\centering

\label{tab:subexpert_cost}

\scalebox{0.65}{
\begin{tabular}{l l c c}
\toprule
\textbf{Model} & \textbf{Dataset} & \textbf{Max Steps} & \textbf{Max Time (hours)} \\
\midrule
Qwen1.5-MoE-2.7B & C4          & 4,200 & 2.09 \\
                 & WikiText-2  & 4,330 & 1.28 \\
                 & OpenWebText & 5,922 & 3.12 \\
\midrule
LLaMA-MoE-3.5B   & C4          & 4,810 & 1.54 \\
                 & WikiText-2  & 5,136 & 1.46 \\
                 & OpenWebText & 4,355 & 3.01 \\
\bottomrule
\end{tabular}
}
\caption{Maximum training steps and wall-clock time among all sub-expert models.
As sub-expert models are trained fully in parallel, we report the maximum value across experts, which determines the overall training time of this stage.}
\label{tab:subexpert_cost}
\end{table}

\section{Gating Network Architecture}
\label{app:gating}

The MoE architecture employed in this work adopts the standard top-$K$ sparsely-gated routing strategy. In this scheme, for each input token, the gating network dynamically selects the $K$ most relevant experts for activation while leaving the rest inactive, thereby maintaining high model capacity with controlled computational cost. During the submodel training phase, only a single expert is assigned per submodel; consequently, the gating network is not utilized, and each submodel is trained as a dense model with a fixed expert. In the subsequent fine-tune stage, the full gating network is introduced and requires only a small number of optimization steps to learn effective and coordinated routing across experts, achieving efficient and stable adaptation without extensive retraining.
\end{document}